\documentclass{article}

\usepackage{microtype}
\usepackage{graphicx}
\usepackage{subcaption}
\usepackage{booktabs} % for professional tables

\usepackage{hyperref}
\usepackage{bm}
\usepackage[dvipsnames]{xcolor}
\usepackage{hyperref}
\usepackage{url}
\usepackage{multirow}
\usepackage{colortbl}
\usepackage{graphicx}

\usepackage{amsmath}
\usepackage{wrapfig}
\usepackage{hyperref}
\usepackage{multirow}
\usepackage{makecell}

\usepackage[accepted]{icml2023} 

\usepackage{amsmath}
\usepackage{amssymb}
\usepackage{mathtools}
\usepackage{amsthm}
\DeclareMathOperator{\E}{E}

\usepackage{lipsum}

\usepackage[capitalize,noabbrev]{cleveref}

\theoremstyle{plain}

\theoremstyle{definition}

\theoremstyle{remark}

\usepackage[textsize=tiny]{todonotes}

\icmltitlerunning{EM-Network: Oracle Guided Self-distillation for Sequence Learning}

\begin{document}

\twocolumn[
\icmltitle{EM-Network: Oracle Guided Self-distillation for Sequence Learning}



\icmlsetsymbol{equal}{*}

\begin{icmlauthorlist}
\icmlauthor{Ji Won Yoon}{yyy}
\icmlauthor{Sunghwan Ahn}{yyy}
\icmlauthor{Hyeonseung Lee}{yyy}
\icmlauthor{Minchan Kim}{yyy}
\icmlauthor{Seok Min Kim}{yyy}
\icmlauthor{Nam Soo Kim}{yyy}
\end{icmlauthorlist}

\icmlaffiliation{yyy}{Department of ECE and INMC, Seoul National University, Seoul,
Republic of Korea}

\icmlcorrespondingauthor{Ji Won Yoon}{jwyoon@hi.snu.ac.kr}
\icmlcorrespondingauthor{Nam Soo Kim}{nkim@snu.ac.kr}

\icmlkeywords{Machine Learning, ICML}

\vskip 0.3in
]



\printAffiliationsAndNotice{}  % leave blank if no need to mention equal contribution

\pdfoutput=1

\begin{abstract}
We introduce EM-Network, a novel self-distillation approach that effectively leverages target information for supervised sequence-to-sequence (seq2seq) learning. In contrast to conventional methods, it is trained with oracle guidance, which is derived from the target sequence. Since the oracle guidance compactly represents the target-side context that can assist the sequence model in solving the task, the EM-Network achieves a better prediction compared to using only the source input. To allow the sequence model to inherit the promising capability of the EM-Network, we propose a new self-distillation strategy, where the original sequence model can benefit from the knowledge of the EM-Network in a one-stage manner. We conduct comprehensive experiments on two types of seq2seq models: connectionist temporal classification (CTC) for speech recognition and attention-based encoder-decoder (AED) for machine translation. Experimental results demonstrate that the EM-Network significantly advances the current state-of-the-art approaches, improving over the best prior work on speech recognition and establishing state-of-the-art performance on WMT'14 and IWSLT'14.
\end{abstract}

\section{Introduction}
\label{Intro}
Throughout the literature on deep learning, sequence-to-sequence (seq2seq) learning has achieved great success in a wide range of applications, especially in speech and natural language processing.
A task of the seq2seq learning is to learn a function $f: \mathcal{X} \rightarrow \mathcal{Y}$ that maps a source sequence $x \in \mathcal{X}$ to a target sequence $y \in \mathcal{Y}$.
Since seq2seq tasks generally suffer from source-target mismatch problems, e.g., unequal length, different modality, and domain mismatch, learning the latent space $\mathcal{Z}$ and how to improve its quality are deemed critically important in sequence modeling.
Examples include various types of source-target alignment \cite{ctc, glowtts, vits}, and feature representation \cite{bert, hubert, data2vec, bert-fuse, bibert}.

However, it is difficult to learn the optimal latent space for the learning task. 
For example, in speech processing, connectionist temporal classification (CTC) \cite{ctc} models often converge to sub-optimal alignment distributions and produce over-confident predictions \cite{ctc-peaky,ctc-peaky2,ctc-peaky3}. 
Since there is an exponential number of possible alignment paths, and the alignment information between source and target sequences is rarely available during training, settling on the optimal alignment $z$ is quite challenging.
% For recent natural language processing (NLP) studies, a huge dataset and pre-trained language model (LM) are highly required to capture the contextualized representation $z$ \cite{bert, bert-fuse, unidrop, bibert}.

For the tasks above, one promising approach is knowledge distillation (KD) \cite{hinton_kd-et-al:scheme} that allows a sequence model (student) to learn a higher-capacity model (teacher)'s latent knowledge $z$, including CTC alignment \cite{tutornet:scheme, guiding_teacher} and hidden representation \cite{bert_1,bert_2, tutornet:scheme}.
Despite their effective regularization in the latent space, such techniques generally rely on a complex teacher model, requiring heavy computational costs.
To mitigate the burden, recent studies have attempted to train the model with a self-distillation setup \cite{rdrop, data2vec, self-distillation1, self-distillation2}, distilling the knowledge within the network itself without any extra teacher network.
In this setting, where the student network acts as its own teacher, there is no guarantee that the distilled knowledge is sufficient to provide an optimal solution for the task.
Also, some self-distillation methods \cite{self-boosting, self-kd, data2vec, self-distillation1, self-distillation2,self-distillation3,self-distillation4,self-distillation5} require an additional update strategy, such as the exponential moving average (EMA) of parameters and noisy student training.

\begin{figure*}[t]
\centering
	\includegraphics[height=4.5cm]{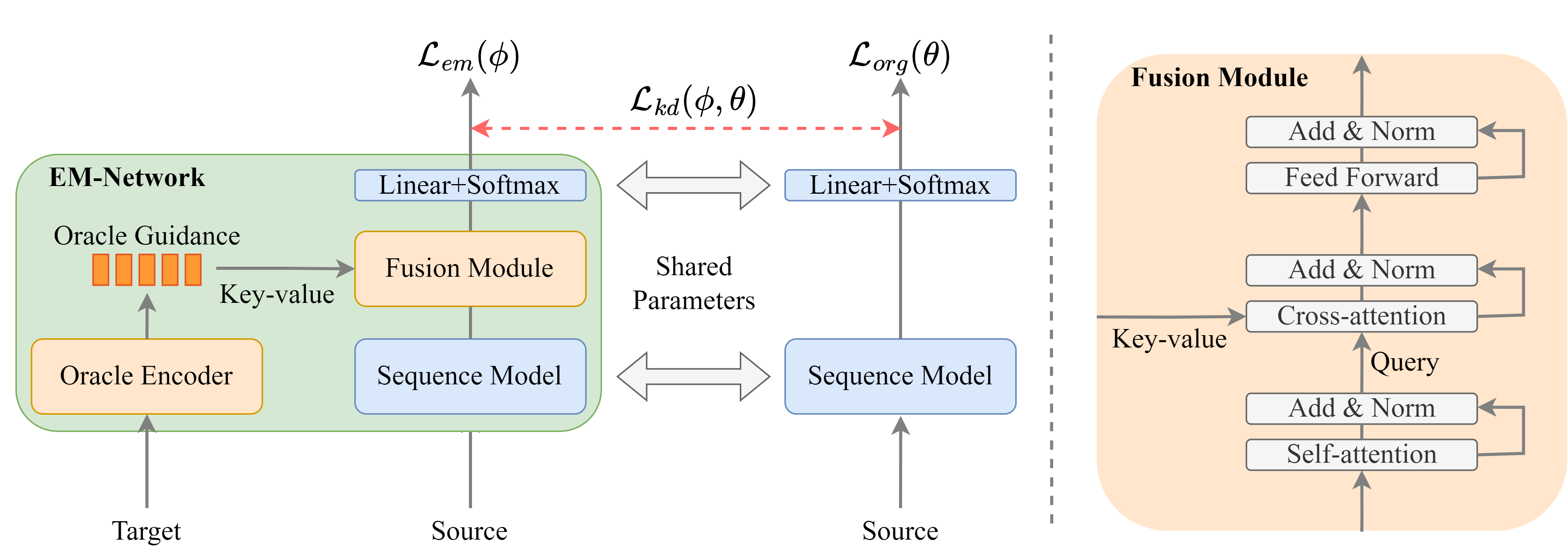}
	\caption{Overview of the EM-Network. Green components ($\phi$) and blue components ($\theta$) represent the parameters of the EM-Network and the sequence model, respectively.
    We can select the architecture of the sequence model depending on the task demands.
    In our experiment for ASR, the sequence model is based on the conventional CTC model. For MT, we apply the autoregressive Transformer \cite{transformer} as the sequence model.
    The oracle encoder consists of an embedding layer and a self-attention-based Transformer encoder layer. In the case of the fusion module, its architecture is based on self and cross-attention layers.
    Specifically, the cross-attention takes the representation of the source as a query and the oracle guidance as key-value pairs.
    }
    % \vspace{-3mm}
	\label{overview}
\end{figure*}

In this paper, we present a novel self-distillation framework, namely EM-Network, that effectively leverages target information for supervised seq2seq learning.
This work departs from previous distillation approaches in three ways.
First, our method leverages the target sequence as the model’s additional training input.  
The proposed target injection enables the EM-Network to generate a higher-quality latent space, resulting in better predictions.
For instance, as shown in Appendix \ref{ctc_align_add}, the EM-Network for CTC produces a more accurate latent alignment $z$ by consuming both acoustic (speech source) and linguistic (text target) embeddings. 
Second, we theoretically demonstrate that the proposed objective function is justified from the Expectation-Maximization (EM)-like algorithm perspective.
Finally, the EM-Network can effectively solve optimization problems for various seq2seq models, ranging from small to large sizes. It is designed to be flexible and can be applied to different architectures and settings, requiring a few additional parameters.
Unlike prior works, our model can be stably optimized without any additional complex update strategies.

The proposed approach involves three components: (I) a standard \emph{sequence model} that performs the original learning task given the source input $x$, (II) an \emph{oracle encoder} that generates the \emph{oracle guidance} from the target $y$, and (III) a \emph{fusion module} that combines the oracle guidance with the sequence model's representation.
The oracle guidance compactly represents the target-side context that can help the sequence model to solve the task.
Thus, given the additional oracle guidance, the EM-Network can make overall improved prediction results compared to the case when it consumes only the source input for training.

Our key insight is to enable the sequence model to inherit the promising capability of the EM-Network.
For this, we design a novel self-distillation strategy, where the sequence model can benefit from the EM-Network's knowledge in a one-stage manner.
The prediction of the EM-Network (teacher mode) serves as a soft label for training the inner sequence model (student mode) that consumes only the source input.
Different from previous self-distillation approaches, the proposed method ensures that the teacher's prediction is inherently more accurate than that of the student since the EM-Network (teacher mode) has access to ground truth during training.
It is important to note that the parameters of the standard sequence model are included in the EM-Network, and only the sequence model is used to generate the prediction during the inference.

The proposed approach is capable of modeling two popular seq2seq networks: CTC and AED.
Even though the two models have different properties in terms of training objective, architecture, and the use of explicit latent alignment, EM-Network can build a general solution with very little task-specific modifications; for the AED model, the masked version of the target $\tilde{y}$ is used instead of the whole target $y$.
Note that our method is designed for supervised learning setting, requiring labeled data.
We conduct comprehensive experiments on multiple benchmarks, including automatic speech recognition (ASR) and machine translation (MT) tasks.
The CTC and AED models are applied to ASR and MT tasks, respectively.
Experimental results demonstrate that the EM-Network improves over the best prior work on ASR and establishes SOTA performance on WMT'14 and IWSLT'14 datasets.

\section{Methodology}
\label{2}
In this section, we first give a high-level overview of the EM-Network and then describe its components in detail.
We further provide a theoretical interpretation of the proposed framework from an EM-like perspective in Section \ref{why}.

\subsection{EM-Network with Oracle Guidance}
As shown in Figure \ref{overview}, we devise the EM-Network as a composition of a standard \emph{sequence model}, an \emph{oracle encoder}, and a \emph{fusion module}.
For the convenience of notation, we let $\phi$ and $\theta$ denote the parameters of the EM-Network and the sequence model, respectively.
Note that the EM-Network includes the parameters of the sequence model, where $\theta \subset \phi$.

To build the proposed model, we start from a sequence model, which maps a source sequence $x \in \mathcal{X}$ to a target sequence $y \in \mathcal{Y}$.
As discussed earlier, such models may not be sufficient enough to obtain the optimal latent space $\mathcal{Z}$ for the task.

To cope with this modeling deficiency, we introduce \emph{oracle guidance} $r \in \mathcal{R}$, which is derived from the target $y$ through the oracle encoder.
Since the oracle guidance has access to the target $y$, it compactly represents the target-side context that can help the sequence model to solve the task.
We combine the oracle guidance $r$ with the sequence model's representation via the fusion module.
The assumption is that given $x$ and $r$, the model can produce better latent space compared to the case when it consumes only the source input $x$.
For example, consider the CTC-based ASR task: the EM-Network, aided by not only the acoustic information (speech source) but also the linguistics (text oracle), is capable of learning the higher-quality latent alignment $z$.
Empirically, we observe that the EM-Network can find a more accurate CTC alignment by leveraging the oracle guidance, as demonstrated in Appendix \ref{ctc_align_add}.

\paragraph{Self-distillation.}
Our key insight is to enable the sequence model to inherit the promising capability of the EM-Network.
For this, the proposed framework is trained in a new self-distillation setup.
The EM-Network first yields the predictions, which serve as soft labels in distillation, by encoding both the source $x$ and target $y$ inputs (teacher mode).
Then, the sequence model learns the parameter $\theta$ to predict the soft labels given the source input $x$ (student mode).
Since the parameters of the standard sequence model are included in the EM-Network, the proposed distillation can be considered as the self-distillation scheme.

\subsection{Training Objective}
In the proposed framework, there are three kinds of losses during the training phase: (1) the original sequence learning loss $\mathcal{L}_{org}(\theta)$, such as CTC loss and frame-wise cross entropy (CE), with the source input $x$ to train the sequence model, (2) the proposed sequence learning loss $\mathcal{L}_{em}(\phi)$ given both source and target inputs for training the EM-Network, and (3) self-distillation loss $\mathcal{L}_{kd}(\phi,\theta)$ that trains the sequence model to predict the soft labels generated from the EM-Network. 
We optimize the following loss to train the proposed framework:
\begin{equation}
   \mathcal{L}_{train}(\phi,\theta) = \mathcal{L}_{org}(\theta) + \mathcal{L}_{em}(\phi)+ \alpha \mathcal{L}_{kd}(\phi,\theta)
   \label{train_loss}
\end{equation}
where $\alpha$ is a tunable parameter.
In our self-distillation manner, the soft labels from the EM-Network are updated in each iteration since $\phi$ is affected by the two losses $\mathcal{L}_{em}(\phi)$ and $\mathcal{L}_{kd}(\phi,\theta)$.
The detailed implementation of $\mathcal{L}_{em}(\phi)$ will be described in Section \ref{imp}.

\paragraph{Inference.}
During the inference, only the sequence model is used to generate the prediction.

\subsection{Implementation Details}
\label{imp}
% \paragraph{Modeling Conditional Probability with Target Input.}
When training the EM-Network with the proposed loss $\mathcal{L}_{em}(\phi)$, the crucial challenge is how to use the target input while preventing the trivial solution, where the model converges with the conditional probability $P(y|x,y) = \delta(y)$.

\paragraph{Objective for CTC-based EM-Network.} 
Before describing the proposed objective $\mathcal{L}_{em}(\phi)$ for the CTC framework, it might be beneficial to briefly discuss the CTC objective function.
Given an input sequence $x$, the core idea of CTC \cite{ctc} is leveraging intermediate alignment $z$ by allowing label repetitions possibly interleaved with a special {\it blank} token ($\epsilon$). 
 CTC trains the sequence model $\theta$ to minimize the following loss function:
\begin{equation}
   \mathcal{L}_{ctc}(\theta) = -\log P(y \vert x; \theta) = -\log\sum_{z \in \mathcal{B}^{-1}(y)} P(z \vert x; \theta)
   \label{ctc}
\end{equation}
where $\mathcal{B}$ is a many-to-one mapping in the CTC algorithm with $y=\mathcal{B}(z)$. $\mathcal{B}^{-1}(y)$ is the set of possible alignments $z$ compatible to $y$.
The mapping is done by merging repeated labels from the paths and then removing all {\it blank} tokens ($\epsilon$), e.g., $\mathcal{B}$(\{$\epsilon$,a,a,$\epsilon$,$\epsilon$,a,b,b\}) = \{aab\}) where \{aab\} denotes $y$ and the sequence \{$\epsilon$,a,a,$\epsilon$,$\epsilon$,a,b,b\} represents $z$.

For the CTC framework, we can easily apply the target $y$ as the EM-Network's input since the CTC algorithm inherently prohibits the trivial solution.
In CTC, the latent alignment $z$ is explicitly adopted, as shown in Eq. (\ref{ctc}).
The relationship between $z$ and $y$ is many-to-one, which can be formulated as $z \in \mathcal{B}^{-1}(y)$.
Intuitively, it is challenging to predict $z$ (many) given the target $y$ (one).
Therefore, the objective function for the CTC-based EM-Network can be given as
\begin{equation}
   \mathcal{L}_{em-ctc}(\phi) = -\log\sum_{z \in \mathcal{B}^{-1}(y)} P(z \vert x, y ; \phi).
   \label{ctc_proposed}
\end{equation}
The CTC-based EM-Network aims at predicting a better alignment $z$ by leveraging the source and target inputs. Therefore, the sequence model distilled from the EM-Network can produce more accurate CTC alignments.

\paragraph{Objective for AED-based EM-Network.} Different from the CTC, the AED model does not explicitly define the latent alignment $z$ and the many-to-one mapping $\mathcal{B}$.
Since simply taking the target $y$ as the AED model's input may cause an obvious but trivial solution, it is difficult to learn a meaningful oracle guidance from the target input $y$.
To sidestep this problem, we employ a masking strategy in the masked language model (MLM) task.
The masked version of the target $\tilde{y}$ is applied as the additional input instead of directly using the target $y$.
Given the target sequence $y$, we use a masking function $\mathcal{M}$, which randomly masks the tokens of $y$ with the probability of $\lambda$. Then, the masked target $\mathcal{M}(y,\lambda)=\tilde{y}$ is fed to the EM-Network as the auxiliary input. The loss function for the AED-based EM-Network can be calculated as follows:
\begin{equation}
   \mathcal{L}_{em-aed}(\phi) = -\log P(y \vert x,\tilde{y}; \phi) \quad \text{where} \ \tilde{y}=\mathcal{M}(y, \lambda).
   \label{ce_em}
\end{equation}
Since the EM-Network for AED can utilize global target features by computing the posterior $P(y|x,\tilde{y};\phi)$, it provides more robust contextualized representations that can benefit the learning task.

Note that the EM-Network has very little task-specific modifications for the AED model; the masked version of the target $\tilde{y}$ is utilized instead of the whole target $y$.

\paragraph{Self-distillation.}
The conventional distillation loss based on CE or Kullback-Leibler (KL)-divergence generally fails to converge in the CTC framework, as reported in previous studies \cite{ctc_kd,ctc_kd2,ctc_kd3, tutornet:scheme}.
Thus, we follow the distillation loss of \citet{tutornet:scheme} for the CTC-based EM-Network, which adopts $l_2$ loss function for transferring the latent alignment.
In the case of the AED case, we utilize the conventional distillation loss, where the KL divergence is computed using softmax outputs between the student and the teacher.

\paragraph{Sequence Model.}
We can flexibly select the architecture of the sequence model depending on the task demands.
In our experiments, we employ various model architectures for the sequence model, including Conformer-CTC \cite{conformer}, Data2vec \cite{data2vec}, Transformer \cite{transformer}, and BiBERT \cite{bibert}.

\paragraph{Oracle Encoder.}
The oracle encoder generates the oracle guidance from the target input. It consists of an embedding layer and a self-attention-based Transformer encoder layer.

\paragraph{Fusion Module.}
The fusion module performs a fusion of the sequence model's representation and the oracle guidance when producing the prediction.
Its architecture is based on self and cross-attention layers.
As depicted in Figure \ref{overview}, the first self-attention layer of the fusion module takes the representation of the sequence model as the input. 
The resulting output from this self-attention layer serves as queries for the subsequent cross-attention layer. 
Furthermore, the cross-attention layer utilizes the oracle guidance $r$ as the key-value pairs.
Through this fusion process, the proposed framework can effectively incorporate the target-side context offered by the oracle guidance $r$, leading to the promising capability of the EM-Network.

\section{Connection to EM Algorithm}
Now we explore the connection with the EM algorithm.
For both CTC and AED frameworks, the EM-Network can be interpreted with a similar perspective.
In this section, we give the theoretical analysis of the CTC-based EM-Network.
The theoretical justification for the AED-based EM-Network are provided in Appendix \ref{derive}.

\label{why}
% In this section, we give theoretical justification on why the proposed framework helps the training of the sequence model.
\paragraph{Standard EM Algorithm.}
The traditional EM algorithm finds maximum likelihood parameters of the model that depends on unobserved latent variables $z$.
It iteratively performs an expectation (E) step and a Maximization step (M step).
The E step defines a Q-function $Q(\theta|\theta^{(t)})$ as the expected value of the log-likelihood of $\theta$, which can be formulated as follows:
\begin{align}
   \quad Q(\theta|\theta^{(t)})=\E_{z|x, \theta^{(t)}}[\log P(x,z;\theta)]. \nonumber
\end{align}%
Then, the M step computes the parameters that maximize the Q-function found on the E step.

\paragraph{EM-like Perspective for EM-Network}
The proposed framework is motivated by the EM algorithm.
In the case of the EM-Network for CTC, the Q-function is calculated as follows:
\begin{align}
   Q(\theta|\phi^{(t)})  &  = \E_{z|x, y, \phi^{(t)}}[\log P(y,z|x;\theta)]\nonumber \\
  &  = \E_{z|x, y, \phi^{(t)}}[\log [\delta(y-\mathcal{B}(z))P(z|x;\theta)]]\nonumber \\
  &  = \sum_{z \in \mathcal{B}^{-1}(y)}P(z|x,y;\phi^{(t)})\log P(z|x;\theta) \nonumber \\
  &  =  - D_{KL}(\underbrace{P(z|x,y;\phi^{(t)})}_{\text{EM-Network}} \Vert \underbrace{P(z|x;\theta)}_{\text{Sequence model}})  \label{em_pers} \\
  & \approx -\mathcal{L}_{kd}(\phi^{(t)}, \theta) \label{kd_em_pers}
\end{align}%
where $P(z|x,y;\phi^{(t)})$ represents the conditional probability of the CTC-based EM-Network in Eq. (\ref{ctc_proposed}), and $\log P(z|x;\theta)$ denotes the CTC-based sequence model in Eq. (\ref{ctc}).
Here, we ignore the constant factor $-H(P(z|x,y;\phi^{(t)}))$ in the fourth equality and the first approximation.
As shown in Eq. (\ref{em_pers}), a negative value of KL-divergence between the EM-Network and the sequence model serves as the Q-function of the EM-Network.
However, as aforementioned, CTC models often fail to converge with the distillation loss using the KL-divergence due to its alignment-free property \cite{ctc_kd, ctc_kd2, ctc_kd3, tutornet:scheme}.
To sidestep this convergence problem, we use $l_{2}$ loss instead of the KL-divergence, corresponding to the distillation loss $\mathcal{L}_{kd}$.
Eq. (\ref{kd_em_pers}) shows that $\mathcal{L}_{kd}$ can be considered as the approximation of the EM-Network's Q-function.

Based on the above perspective, we can actually show our proposed training objective is a lower bound for the log-likelihood of the original sequence model.
Our goal is to maximize the log-likelihood below,
\begin{align}
\label{lower_ctc}
  \log P(y|x;\theta) 
  & = \log\sum_{z}\delta(y-\mathcal{B}(z))P(z|x;\theta)  \nonumber \\
  % \frac{P(z|x,y;\phi^{(t)})}{P(z|x,y;\phi^{(t)})} \nonumber \\
  & = \log\sum_{z \in \mathcal{B}^{-1}(y)}P(z|x;\theta)\frac{P(z|x,y;\phi^{(t)})}{P(z|x,y;\phi^{(t)})} \nonumber \\
  & = \log\sum_{z \in \mathcal{B}^{-1}(y)}P(z|x,y;\phi^{(t)})\frac{P(z|x;\theta)}{P(z|x,y;\phi^{(t)})} \nonumber \\
  & \geq \sum_{z \in \mathcal{B}^{-1}(y)}P(z|x,y;\phi^{(t)})\log\frac{P(z|x;\theta)}{P(z|x,y;\phi^{(t)})} \nonumber \\
    &  =  - D_{KL}(P(z|x,y;\phi^{(t)}) \Vert P(z|x;\theta)) \nonumber \\
  & \approx -\mathcal{L}_{kd}(\phi^{(t)}, \theta) 
  \end{align}
  
where the first inequality follows from Jensen’s inequality. From Eq. (\ref{lower_ctc}), we can confirm that a negative value of $\mathcal{L}_{kd}$ serves as the lower bound for the log-likelihood of the original CTC-based sequence model $\log P(y|x;\theta)$.
Therefore, both Eq. (\ref{kd_em_pers}) and  Eq. (\ref{lower_ctc}) indicate that maximizing the EM-Network's Q-function is equivalent to maximizing the lower bound of the sequence model's likelihood.
The tight lower bound will be additionally discussed in Appendix \ref{ctc_align_add}.

\begin{table*}[t]
\caption{(\textbf{Speech recognition}) Comparison of word error rate (\%) on LibriSpeech test dataset. Bold represents superior results.}
\vskip 0.15in
\centering
\begin{small}
\begin{tabular}{lccccccc}
\hline
\multirow{2}{*}{Baseline Model}& \multicolumn{3}{c}{\multirow{2}{*}{\# of Params.}} & \multicolumn{2}{c}{w/o LM} & \multicolumn{2}{c}{w/ LM}\\
 &  & && clean & other & clean & other\\
\hline
Conformer-CTC Large & \multicolumn{3}{c}{122 M} & 2.78 & 6.18 & 2.01 &	4.69\\
Conformer-CTC Small & \multicolumn{3}{c}{13 M} & 4.87 & 12.05 & 2.85 &	8.34\\
\hline
\multirow{2}{*}{Method} & \multirow{2}{*}{\begin{tabular}[c]{@{}c@{}}Additional \# of \\Params. for Teacher\end{tabular}} & \multirow{2}{*}{Student} & \multirow{2}{*}{KD} & \multicolumn{2}{c}{w/o LM} & \multicolumn{2}{c}{w/ LM}\\
 &  & && clean & other & clean & other\\
\hline
Baseline & None & \multirow{4}{*}{{\begin{tabular}[c]{@{}c@{}}Conformer-CTC \\ Small\end{tabular}}} & None & 4.87 & 12.05 & 2.85	& 8.34 \\
Guided CTC training &122 M &  & Offline & 4.63 & 11.56 & 2.88 &	8.27\\
SKD & 122 M &  & Offline & 4.53 & 11.33 & 2.82 &	8.00 \\
\textbf{Ours, EM-Network} & 1 M &  & \textbf{Self} & \textbf{4.29} & \textbf{10.81} & 	\textbf{2.70} &	\textbf{7.80}\\
\hline
\end{tabular}
\end{small}
\label{superv}
\end{table*}

\begin{table}[t]
\caption{(\textbf{Speech recognition}) Word error rate (\%) on LibriSpeech test set with greedy decoding. The results of ILS-SSL and HuBERT were implemented using the public fairseq \cite{fairseq} toolkit. For data2vec and wav2vec 2.0 models, we evaluated the performance using the checkpoints provided by the fairseq.}
\vskip 0.15in
\centering
\begin{small}
\begin{tabular}{lcc}
\hline
Method & \multicolumn{1}{l}{clean} & \multicolumn{1}{l}{other} \\ \hline
HuBERT Base  & 3.79 & 9.04\\
wav2vec 2.0 Base &3.40&8.42\\
ILS-SSL Base  & 3.44 & 7.79 \\
data2vec Base  & 2.78 & 7.02\\
\textbf{Ours, EM-Network}&   \textbf{2.66}&  \textbf{6.72} \\ 
\hline
\label{ctc_result}
\end{tabular}
\end{small}
% \vspace{-3mm}
\end{table}

Maximizing the lower bound (minimizing $\mathcal{L}_{kd}$) is also closely related to the upper bound for the sequence model's log-likelihood. 
In conventional KD, it is generally assumed that the teacher's performance determines the upper bound for the student \cite{upper1,upper2,upper3}. 
Under the assumption, the log-likelihood of the EM-Network (teacher mode) can be considered as the upper bound for the log-likelihood of the sequence model.
Different from the previous offline KD methods, where the teacher model is typically fixed, we update the EM-Network and sequence model simultaneously, as shown in Figure \ref{loss}.
The proposed distillation loss $\mathcal{L}_{kd}$ regularizes the EM-Network (teacher mode) to learn the knowledge of the sequence model (student mode), providing a tight upper bound for the sequence model.
For the counterpart, minimizing $\mathcal{L}_{kd}$ also corresponds to making the sequence model mimic the behavior of the EM-Network.
This implies that the log-likelihood of the sequence model is close to its upper bound.
Thus, maximizing the lower bound allows us to derive the tight upper bound.
Considering that the upper bound can be maximized by minimizing the training loss $\mathcal{L}_{em}$ of the EM-Network, the proposed objective function partially maximizes the likelihood of the sequence model.

\section{Experimental Setup}
\label{ex_set}

\paragraph{Speech Recognition.}
For the fully supervised setting, ASR models were implemented in the NeMo \cite{nemo:scheme} toolkit.
We trained the EM-Network with the full 960 hours of LibriSpeech (LS-960) \cite{librispeech}, and the sequence model ($\theta$) was based on a conformer-CTC architecture consisting of 16 conformer \cite{conformer} blocks with 176 dimensions.
In the case of the language model (LM), we used the Transformer LM.
When applying our approach to the fine-tuning stage of self-supervised learning (SSL), experiments were conducted using the fairseq \cite{fairseq} toolkit. 
We first pre-trained the sequence model ($\theta$) on LS-960 by using the objective of the data2vec since it is the current SOTA model in the SSL literature.
Then, the whole EM-Network ($\phi$) was fine-tuned with LS-960.
The architecture of the sequence model followed the Base setup of the conventional SSL-based model, consisting of 12 Transformer blocks.
The tunable parameter $\alpha$ was experimentally set to 2.
The detailed implementation is additionally presented in Appendix \ref{imple_detail}.

\paragraph{Machine Translation.}
We evaluated the EM-Network on IWSLT'14 and WMT'14 datasets for English-to-German (En-De) and German-to-English (De-En) translation tasks.
The implementation of the EM-Network was based on the official source code provided by the previous work \cite{bibert}.
For the fully supervised setting, the sequence model was based on the autoregressive Transformer model \cite{transformer}, and the parameter $\alpha$ was set to 5.
In the case of the pre-training, the sequence model was trained on the same English and German texts as BiBERT \cite{bibert}, derived from the OSCAR \cite{oscar} corpus.
When applying our work to the SSL's fine-tuning phase, the sequence model's architecture was based on the BiBERT, a recent SOTA method on the MT task.
During the fine-tuning, the parameter $\alpha$ and the masking ratio $\lambda$ were set to 2 and 50 \%, respectively.

\section{Main Empirical results}
\label{ex_result}
To evaluate our approach and show its effectiveness, we conducted comprehensive experiments on ASR and MT tasks.
Firstly, our approach was compared with other distillation methods in a fully supervised setting.
Also, since research interest in SSL models has grown rapidly, we applied our algorithm to the fine-tuning task of SSL.

\begin{table*}[h]
\caption{(\textbf{Machine translation}) Comparison of BLEU score on IWSLT'14 test set. Bold represents superior results.}
\vskip 0.15in
\centering
\begin{small}
\begin{tabular}{lccccc}
\hline
Baseline Model & \multicolumn{3}{c}{\# of Params.} & En-De & De-En \\
\hline
BiBERT & \multicolumn{3}{c}{206 M} & 30.48 & 38.66 \\
Transformer & \multicolumn{3}{c}{120 M} & 27.87 & 34.18 \\
\hline
Method & \begin{tabular}[c]{@{}c@{}}Additional \# of \\Params. for Teacher\end{tabular} & Student & KD & En-De & De-En \\
\hline
Baseline & None & \multirow{5}{*}{Transformer} & None & 27.87 & 34.18 \\
Standard KD & 206 M & & Offline & 29.00	& 35.46 \\
Sequence-level KD & 206 M &  & Offline & 29.33 & 35.97 \\
R-Drop & 0 M &  & Self & 29.43 & 35.99 \\
\textbf{Ours, EM-Network} & 19 M &  & \textbf{Self} & \textbf{29.94} & \textbf{36.53} \\
\hline
\end{tabular}
\end{small}
\label{superv_mt}
\end{table*}

\begin{table}[t]
\caption{(\textbf{Machine translation}) Comparison of our EM-Network and most recent existing methods on IWSLT’14 and WMT'14 datasets. We reimplemented the results of BiBERT using the public code \cite{bibert},
which are shown inside the parentheses.}
\vskip 0.15in
\centering
\begin{small}
\begin{subtable}[h]{0.45\textwidth}
\centering
\caption{BLEU score on IWSLT’14 test set.}
\begin{tabular}{lcc}
\hline
Method                    & En-De & De-En          \\
\hline
Adversarial MLE           & -     & 35.18          \\
DynamicConv               & -     & 35.20          \\
Macaron Net               & -     & 35.40          \\
BERT-Fuse                 & 30.45 & 36.11          \\
MAT                       & -     & 36.22          \\
Mixed Representations     & 29.93 & 36.41          \\
UniDrop                   & 29.99 & 36.88          \\
\multirow{2}{*}{BiBERT} &30.45 & 38.61\\
                  & (30.48) & (38.66)          \\
\textbf{Ours, EM-Network} &   \textbf{31.80}    & \textbf{39.49} \\
\hline
\label{iwslt_result}
\end{tabular}
\end{subtable}
\hfill
\begin{subtable}[h]{0.45\textwidth}
\centering
\caption{BLEU score on WMT'14 newstest2014 test set.}
\begin{tabular}{lcc}
\hline
Method                    & En-De & De-En          \\
\hline
Large Batch Training &  29.3     & -          \\
Evolved Transformer       & 29.8     &  -         \\
BERT Init. (12 layers) & 30.6 &  33.6         \\
BERT-Fuse                 & 30.75 &  -        \\
\multirow{2}{*}{BIBERT} & 31.26 & 34.94 \\
                  & (30.80) &  (34.53)        \\
\textbf{Ours, EM-Network} & \textbf{31.30}     & \textbf{35.40} \ \\
\hline
\label{wmt_result}
\end{tabular}
\end{subtable}
\end{small}
\label{mt_result}
% \vspace{-3mm}
\end{table}

\paragraph{Speech Recognition.} 
Table \ref{superv} shows the word error rate (WER) results on the LibriSpeech test set.
We applied Guided CTC training \cite{guided} and SKD \cite{tutornet:scheme} as the competing KD methods for ASR.
The Conformer-CTC Large was adopted as the teacher for Guided CTC training and SKD.
The Guided CTC training is the effective KD method for CTC-based ASR model. The student can be guided to align with the frame-level alignment of the teacher by using the guided mask.
The SKD is the recent and promising KD method in the ASR task, which transfers the frame-level posterior of the teacher CTC network.
As shown in Table \ref{superv}, compared to the conventional KD methods using the SOTA ASR model as the teacher, EM-Network showed the best performance in all configurations.
With the greedy decoding, the EM-Network yielded WER 4.29 \% on test-clean while achieving a relative error rate reduction (RERR) of 11.91 \%.
When applying beam-search decoding with LM, it achieved WER 2.70 \%/7.80 \% and RERR 5.26 \%/6.47 \% on the clean/other datasets.
Even though both Guided CTC training and SKD required a complex teacher model (122 M parameters) with heavy computational costs, only 1 M parameters were required to distill the knowledge in the proposed framework.

To mostly show the effectiveness, we took the super strong pre-trained data2vec \cite{data2vec} model as our backbone and fine-tuned it using our method.
We compared the EM-Network with previous SOTA works from the literature, including wav2vec 2.0 \cite{wav2vec}, HuBERT \cite{hubert}, ILS-SSL \cite{ils-ssl}, and data2vec \cite{data2vec}.
As shown in Table \ref{ctc_result}, EM-Network outperformed the recent approaches on the LibriSpeech.
On the LibriSpeech, we improved upon the best data2vec by 0.12 \%/0.3 \% on the clean/other test datasets with greedy decoding, yielding a RERR of 4.32 \%/4.27 \% compared to the data2vec baseline.
Considering that the data2vec already produced promising ASR results, the EM-Network yielded significant improvements.
The EM-Network, aided by the oracle guidance, was capable of learning better latent alignment by using both source and target embeddings.
Thus, the sequence model could produce strong WER performance even without the LM.

\paragraph{Machine Translation.}
For the MT task, we firstly compared the EM-Network with the currently popular distillation methods, such as standard KD \cite{hinton_kd-et-al:scheme}, sequence-level KD \cite{seq-kd} and R-Drop \cite{rdrop}. The sequence-level KD is a widely-used KD method in sequence generation tasks, and the R-Drop is a recent and effective regularization method built upon dropout. As the teacher model for the sequence-level KD, we adopted BiBERT \cite{bibert}, a SOTA approach for the MT task. 
The results in Table \ref{superv_mt} report that the EM-Network yielded the best BLEU performance on IWSLT'14 dataset, yielding 29.94 and 36.53 on En-De and De-En translation tasks, respectively.

The main difference from the R-Drop was the training loss $\mathcal{L}_{em}$ of the EM-Network. In contrast to the R-Drop that only consumed the source input, the proposed method could utilize the oracle guidance derived from the target input and model interactions within
the target space.
From the results, we can verify that the oracle guidance effectively provided the global target-side information while producing a considerable performance gain.
Compared to the R-Drop, which did not require additional parameters, 19 M parameters were used to train the EM-Network.
However, since the auxiliary modules, including the oracle encoder and fusion module, were removed during the inference, the additional computational load was only required for the training procedure.
Considering the performance improvements, this computational burden during the training seemed reasonable.
Also, the additional parameters would be much smaller when we reduced the vocabulary size.

In addition to the previous experiments, we proceeded to verify whether the EM-Network performed well for the fine-tuning task of SSL.
We chose the recent strong and popular BiBERT \cite{bibert} model as our backbone.
Table \ref{iwslt_result} shows a comparison of the proposed method with the recent approaches on IWSLT'14 En-De and De-En translations.
We chose the eight best-performed MT algorithms to date: 1) Adversarial MLE \cite{adversarial_mle}, 2) DynamicConv \cite{dynamic_conv} 3) Macaron Net \cite{macaronnet}, 4) BERT-Fuse \cite{bert-fuse}, 5) multi-branch encoders (MAT) \cite{mat}, 6) mixed representations from different tokenizers (Mixed Representation) \cite{mixed_representation}, 7) uniting different dropout techniques (UniDrop) \cite{unidrop}, and 8) BiBERT \cite{bibert}.
From the results, it is verified that EM-Network outperformed all of them and yielded around 1.4 (En-De) and 0.9 (De-En) BLEU point gains over the previous SOTA results.

For the high-resource scenario, we conducted experiments on WMT'14 En-De and De-En translations and compared the EM-Network with prior existing works that achieved high BLEU scores on WMT'14 dataset, including large batch training \cite{trans+large}, Evolved Transformer \cite{evovled_trans}, initializing the BERT by leveraging pre-trained checkpoints \cite{bert_init}, BERT-Fuse \cite{bert-fuse}, and BiBERT \cite{bibert}.
From Table \ref{wmt_result}, we verified that our model also gave the best BLEU scores on the high-resource WMT'14 En-De and De-En translations.

\section{Analysis}
\label{analysis_sec}
\paragraph{Training Loss Curves.}
\begin{figure}[t]
% \vskip 0.15in
\centering
	\includegraphics[height=4.3cm]{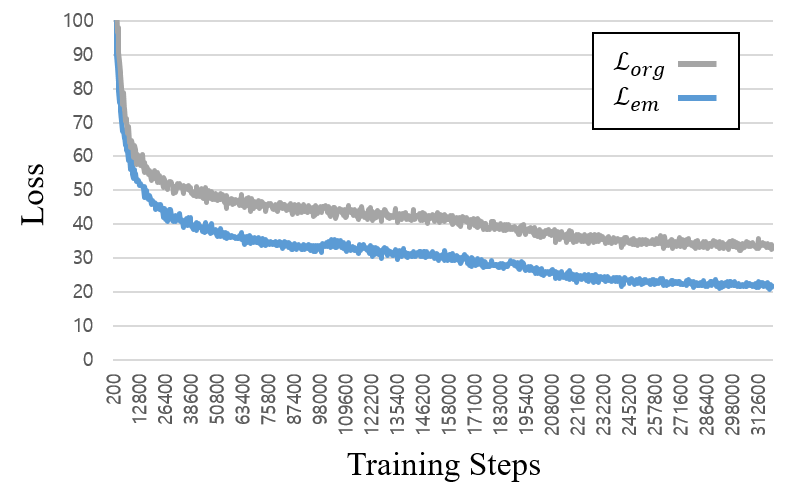}
	\caption{Training loss curves of EM-Network for ASR task.
    }
    % \vspace{-3mm}
	\label{loss}
\end{figure}

\begin{figure*}[t]
\centering
\includegraphics[height=5cm]{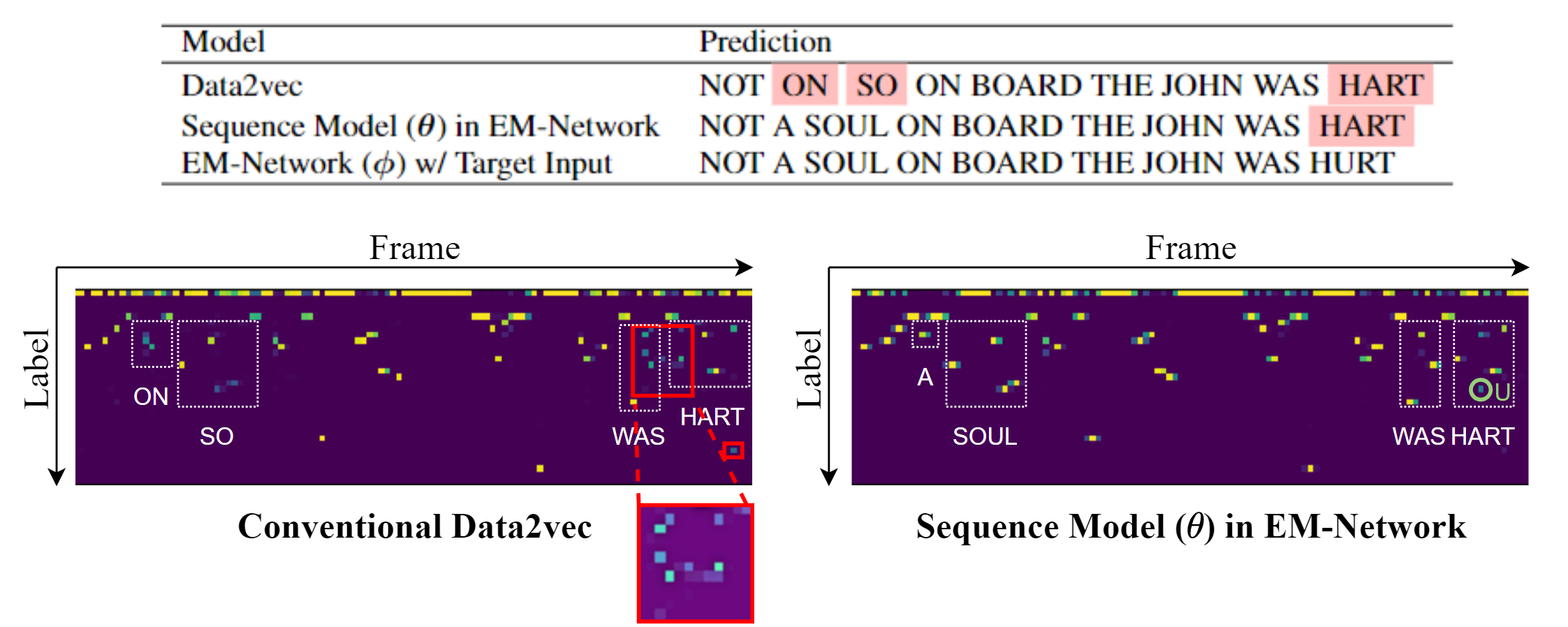}
\caption{Total frame-wise softmax output examples in test-other dataset, where the target reference is ``NOT A SOUL ON BOARD THE JOHN WAS HURT".
The x-axis refers to acoustic frames, and the y-axis refers to
the character labels. The first label index denotes the “blank" label.}
\label{align}
\end{figure*}

Figure \ref{loss} shows the training loss curves ($\mathcal{L}_{org}$ and $\mathcal{L}_{em}$) of the EM-Network for the ASR task.
We observed that the loss $\mathcal{L}_{em}$ of the whole EM-Network ($\phi$) consistently decreased, implying that soft labels from the EM-Network were updated in each iteration.
In this way, unlike the conventional distillation that the knowledge is generally fixed, the EM-Network transferred the knowledge more adaptively.
As shown in Figure \ref{loss}, $\mathcal{L}_{em}$ was lower than $\mathcal{L}_{org}$ during the training.
This means that the prediction of the EM-Network (teacher mode) was more accurate than that of the inner sequence model (student mode). Thus, the sequence model could effectively benefit from the soft labels of the EM-Network. 
Also, compared to previous self-distillation methods \cite{self-boosting,self-kd,data2vec, self-distillation1, self-distillation2,self-distillation3,self-distillation4,self-distillation5}, the proposed distillation framework could be stably optimized by simply applying the additional losses $\mathcal{L}_{em}$ and $\mathcal{L}_{kd}$.

\paragraph{CTC Alignment.}
As visualized in Figure \ref{align}, we contrasted the total frame-wise softmax outputs of the best baseline data2vec and our sequence model $\theta$ in the EM-Network.
The argmax value of the frame-wise label probability corresponds to the predicted CTC alignment.
In Figure \ref{align}, the conventional data2vec converted a given speech into ``NOT ON SO ON BOARD THE JOHN WAS HART" and made erroneous predictions with “ON SO" and “HART". 
When considering only the acoustic feature (speech voice), it is challenging to distinguish “ON SO"/“A SOUL" and “HART"/“HURT".
However, the proposed sequence model provided a more accurate prediction than the data2vec.
Different from the previous approaches, the EM-Network performed a fusion of the acoustic (speech source input) and linguistic (text target input) representations.
Since the EM-Network could consider not only the acoustic information but also the linguistic one, the sequence model learned better alignment and obtained satisfactory performance.
Even though there was a single erroneous word “HART", we can discover the probability regarding "U" (green circle in Figure \ref{align}) in the prediction of the sequence model.
Also, unlike the data2vec that included many irrelevant label probabilities (red boxes in Figure \ref{align}), our sequence model had relatively fewer redundant ones, implying that the oracle guidance effectively reduced the candidates of the latent CTC alignment $z$.
From the results, it is verified that the proposed framework could learn the promising CTC alignment by using the oracle guidance and the self-distillation setup.

\paragraph{Translation Consistency.}

The token repetition ratio represents the degree of the inconsistency problem in MT \cite{alignart,align2}.
We computed the token repetition ratio by dividing all consecutively repeating tokens by the total number of tokens.
The teacher forcing and the greedy decoding were applied to fairly compare the predictions.
From the results in Table \ref{token_rep}, we can confirm that the EM-Network effectively addressed the token repetition issue compared to the prior SOTA approach. 
In the case of the conventional BiBERT, it only considered the past target information during the training via the teacher forcing technique.
However, the EM-Network leveraged a more global target context by using the target input $\tilde{y}$.
Since our approach used both past and future target information in generating the prediction, it alleviated translation inconsistency and produced better quality.
In other words, the sequence model in the proposed framework could benefit from a more optimal and consistent representation of the EM-Network.

\begin{table}[t]
\caption{Token repetition ratio on IWSLT'14 test dataset.}
\vskip 0.15in
\centering
\begin{small}
\begin{tabular}{lcc}
\hline
Model          & \multicolumn{1}{c}{En-De} & De-En\\ \hline
Reference Test Set           & \multicolumn{1}{c}{0.23}      &  0.18      \\ \hline
BiBERT                       & \multicolumn{1}{c}{4.76}  &   5.62   \\ 
Sequence Model (\bm{$\theta$}) in EM-Network & \multicolumn{1}{c}{4.57}  &  4.88       \\ 
EM-Network (\bm{$\phi$}) w/ Target Input  & \multicolumn{1}{c}{4.42}  &  4.70    \\ \hline
\end{tabular}
\label{token_rep}
% \vspace{-5mm}
\end{small}
\end{table}

\paragraph{Cross-attention Visualization.}
\begin{figure}[t]
% \vskip 0.15in
\centering
	\includegraphics[height=4cm]{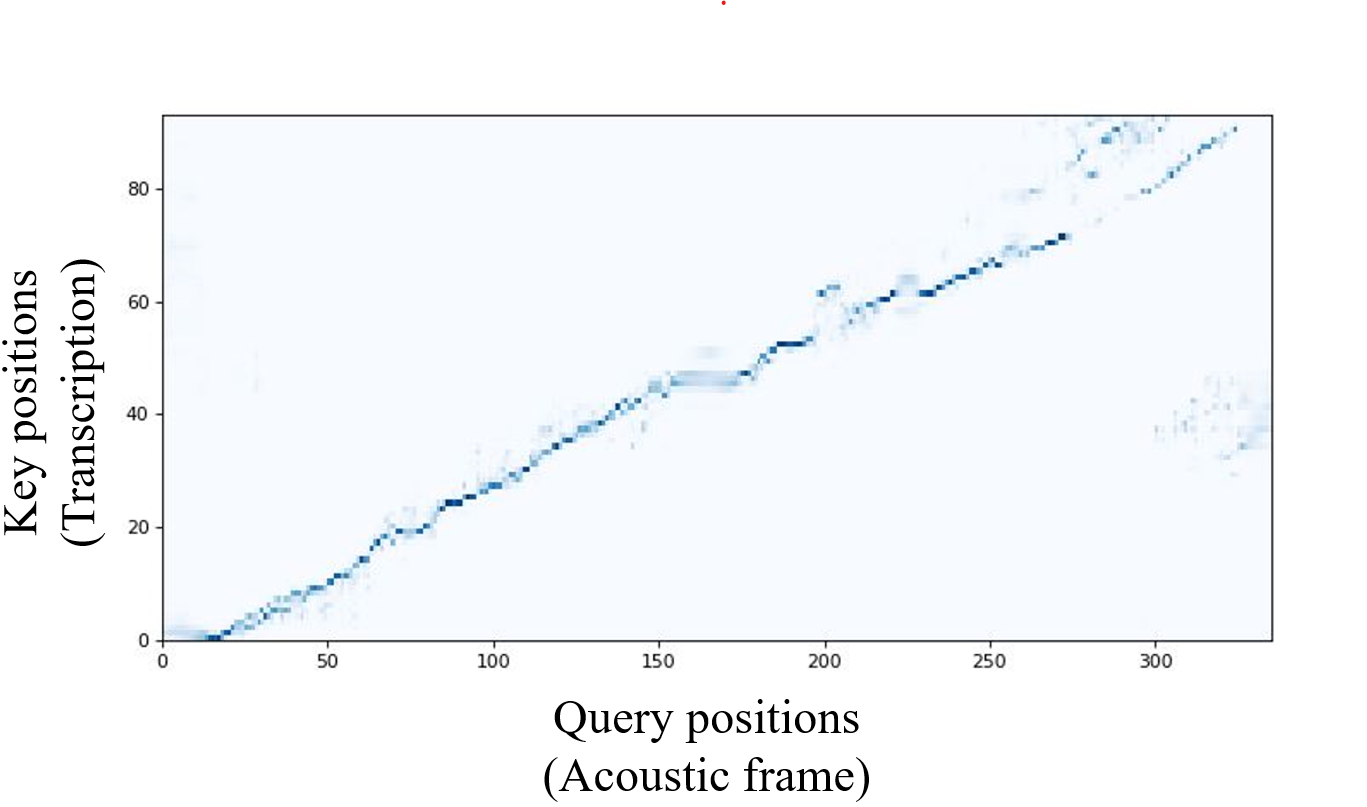}
	\caption{Cross-attention scores of the fusion module.
    }
    % \vspace{-3mm}
	\label{attention}
\end{figure}
As shown in Figure \ref{overview}, the fusion module was based on the self and cross-attention layers.
In the cross-attention layer, the representation of the source served as the query, while the oracle guidance was employed as the key-value pairs.
In Figure \ref{attention}, we visualize the cross-attention scores of the fusion module to confirm that the EM-Network was properly trained using both $x$ and $y$ inputs. The x-axis refers to acoustic frames, and the y-axis refers to transcription. If the EM-Network simply copied the target information, the attention would have an almost diagonal alignment along the key position (text), disregarding the length of the query (acoustic frame).
However, as shown in Figure \ref{attention}, the attention scores were correctly computed along with the acoustic frames (query).
This implies that the EM-Network considered both source (acoustic) and target (linguistic) inputs during the training, preventing the aforementioned trivial solution.

\section{Related Works}

\paragraph{Deep Mutual Learning.}
The proposed self-distillation acted as a regularization for the seq2seq model, like a deep mutual learning scheme \cite{deepmutual}, where the students learn collaboratively and teach each other. However, the main difference from the previous mutual learning approaches \cite{deepmutual, rdrop} was that the proposed method utilized the oracle guidance instead of merely considering the ground truth as the sole target in training.
For the conventional mutual learning framework, the model with teacher mode had a similar loss value to its student.
However, our approach ensured that the prediction of the EM-Network (teacher mode), aided by the oracle guidance, was more accurate than that of the inner sequence model (student mode), as shown in Figure \ref{loss}.
This implies that the EM-Network was sufficiently supportive in transferring the knowledge to the sequence model.
In addition, we attempted to apply the proposed self-distillation to both AED and CTC frameworks in a unified way, an unexplored area in mutual learning research.

\paragraph{Training with Target Input.} 
There have been many techniques in the seq2seq learning literature that exploit the target-related information.
Autoregressive approaches \cite{transformer, chorowski-et-al:scheme, graves-rnn-t:scheme} have adopted a teacher forcing technique that supplies a ground truth value as conditional input during the training.
To make good global planning, \citet{guiding_teacher} proposed the additional seer decoder into the encoder-decoder framework, which embodies future ground truth to guide the behaviors of the conventional decoder.
In the case of the non-autoregressive framework, some studies \cite{mask-ctc,imputer} attempted to improve the CTC model's prediction by conditioning on previously generated tokens.
Our motivation was quite different from the conventional methods.
Inspired by the EM algorithm, the proposed method aimed at learning the promising latent space by leveraging the oracle guidance.
Also, we went much further regarding the experiments' scale and analysis depth. 
Firstly, the proposed method was capable of modeling both CTC and AED models so that it could be applied to a wide range of tasks.
We showed its effectiveness for the fully supervised setting and the SSL's fine-tuning task.
We also investigated how the proposed training objective can improve the sequence model from the EM-like perspective.

\paragraph{Text Injection ASR.} 
Recently, several studies have attempted to learn shared speech and text representation jointly \cite{speechlm, tessp, token2vec, text-injection1, text-injection2}, especially for self-supervised learning scenarios.
In contrast to the previous techniques, our approach focused on injecting text information within a distillation framework, where the teacher model had access to additional text embeddings.
The proposed target injection provided the teacher model with enriched textual understanding, playing a crucial role in providing more optimal guidance to the student.
By incorporating this text injection methodology, we aimed to enhance the distillation framework and effectively solve optimization problems for supervised seq2seq learning.

\section{Conclusions}
In this paper, we proposed a novel distillation approach, namely EM-Network, that can effectively utilize the target information for seq2seq learning.
Instead of merely considering the ground truth as the sole target in training, EM-Network was trained with the oracle guidance derived from the target.
The proposed self-distillation framework enabled the sequence model to benefit from the EM-Network's knowledge in a one-stage manner.
We theoretically showed that our training objective could be justified from the EM-like perspective.
Empirically, EM-Network significantly advanced the current state-of-the-art approaches on ASR and MT tasks. 
We hope our study will draw more attention from the community toward a richer view of the distillation for seq2seq learning.

\newpage
\bibliography{example_paper}
\bibliographystyle{icml2023}

%%%%%%%%%%%%%%%%%%%%%%%%%%%%%%%%%%%%%%%%%%%%%%%%%%%%%%%%%%%%%%%%%%%%%%%%%%%%%%%
%%%%%%%%%%%%%%%%%%%%%%%%%%%%%%%%%%%%%%%%%%%%%%%%%%%%%%%%%%%%%%%%%%%%%%%%%%%%%%%
% APPENDIX
%%%%%%%%%%%%%%%%%%%%%%%%%%%%%%%%%%%%%%%%%%%%%%%%%%%%%%%%%%%%%%%%%%%%%%%%%%%%%%%
%%%%%%%%%%%%%%%%%%%%%%%%%%%%%%%%%%%%%%%%%%%%%%%%%%%%%%%%%%%%%%%%%%%%%%%%%%%%%%%
\newpage
\appendix
\onecolumn
\section{Source Code}
Source code can be found at \url{https://github.com/Yoon-Project/EM-Network-official}.

\section{Derivation for AED-based EM-Network}
\label{derive}

\begin{align}
  \log P(y|x;\theta) 
  & = \log\sum_{\tilde{y}}\delta(\tilde{y}-\mathcal{M}(y,\lambda))P(y|x;\theta)\frac{P(y|x,\tilde{y};\phi^{(t)})}{P(y|x,\tilde{y};\phi^{(t)})} \nonumber \\
  & = \log\sum_{\tilde{y}=\mathcal{M}(y,\lambda)}P(y|x;\theta)\frac{P(y|x,\tilde{y};\phi^{(t)})}{P(y|x,\tilde{y};\phi^{(t)})} \nonumber \\
  & = \log\sum_{\tilde{y}=\mathcal{M}(y,\lambda)}P(y|x,\tilde{y};\phi^{(t)})\frac{P(y|x;\theta)}{P(y|x,\tilde{y};\phi^{(t)})} \nonumber \\
  & \geq \sum_{\tilde{y}=\mathcal{M}(y,\lambda)}P(y|x,\tilde{y};\phi^{(t)})\log\frac{P(y|x;\theta)}{P(y|x,\tilde{y};\phi^{(t)})} \nonumber \\
  & = - D_{KL}(P(y|x,\tilde{y};\phi^{(t)}) \Vert P(y|x;\theta)) \nonumber \\
  & = - \mathcal{L}_{kd}(\phi^{(t)}, \theta) \nonumber
  \end{align}

\section{Implementation Details}
\label{imple_detail}
\begin{table*}[h]
\caption{The number of parameters for each model where $\theta$ represents the parameters of the sequence model and $\rho$ denotes the parameters of the other modules, including the oracle encoder and fusion module.}
\vskip 0.15in
\centering
\begin{small}
\begin{tabular}{ccc}
\hline
Task & Method & \# of Param. \\
\hline
\multirow{2}{*}{ASR} &data2vec Base  & 93 M\\
&EM-Network& 103 M ($\theta$: 93 M, $\rho$: 10 M) \\ 
\hline
\multirow{2}{*}{MT (IWSLT'14)} & BiBERT  & 206 M \\
&EM-Network& 225 M ($\theta$: 206 M, $\rho$: 19 M) \\ 
\hline
\multirow{2}{*}{MT (WMT'14)} & BiBERT  & 324 M\\
&EM-Network& 374 M ($\theta$: 324 M, $\rho$: 50 M) \\ 
\hline
\label{param}
\end{tabular}
\end{small}
\end{table*}
\paragraph{Speech Recognition.}

For the fully-supervised setting, we used four Quadro RTX 8000 GPUs (each with 48GB of memory), and 100 epochs were spent for training the models.
AdamW algorithm \cite{adamw} was employed as an optimizer with an initial learning rate of 5.0.
We used SpecAugment \cite{specaugment} for training the Conformer model. The number of frequency and time masks were set to 2 and 5, respectively. The width of the frequency and time were set to 27 and 0.05.
To preprocess the data for the Conformer model, we used byte-pair-encoding (BPE) tokens with a vocabulary size of 128. 
In the case of the stride of conv block, we followed the configuration provided by the NeMo toolkit.
In the case of Conformer-CTC large, the current SOTA ASR model, we used the pre-trained checkpoint provided by the NeMo \cite{nemo:scheme} toolkit.

For the pre-training stage, we followed the pre-training regime of data2vec \cite{data2vec}.
After the pre-training, pre-trained models were fine-tuned for the ASR task by applying a linear projection layer.
We fine-tuned the model with Adam \cite{adam} optimizer with an initial learning rate of 1e-4.
For the LibriSpeech, the character set had a total of 29 labels plus a word boundary token.
The BASE setup contained 12 Transformer blocks with model dimension 768 and 8 attention heads.
Also, BASE models utilized a batch size of 3.2m samples per GPU and were fine-tuned on eight Quadro RTX 8000 GPUs (each with 48GB of memory), where training updates were set to 320k.

\paragraph{Machine Translation.}
Regarding the selection of the best checkpoint, we selected one best checkpoint based upon the validation loss, following the BiBERT \cite{bibert}. 
For the IWSLT'14 dataset, we used four Titan V GPUs (each with 12GB of memory) with 2048 tokens per GPU and accumulated the gradient 4 times. The optimizer was Adam \cite{adam}, with a learning rate of 0.0004.
We used a temperature of 1.0 for the soft-label distillation.
We followed the same byte-pair encoding (BPE) settings of the previous study \cite{bibert}.
At inference time, we applied beam search with width 4 and a length penalty of 0.6.
In the case of the WMT'14 dataset, newstest2012 and newstest2013 were combined as the validation set, and we used newstest2014 as the test set. 
Followed by the BiBERT\cite{bibert}, we used a unified 52K vocabulary for the decoder.
We utilized four Quadro RTX 8000 GPUs (each with 48GB of memory) with a batch size of 4096 tokens per GPU. The gradient was accumulated 32 times. The initial learning rate was set to 0.001.

\paragraph{Number of Parameter.}
Table \ref{param} compares the number of parameters between the best seq2seq baseline and the EM-Network. 
For the convenience of notation,
we let $\rho$ denotes the parameters of the additional modules, including the oracle encoder and fusion module.
We used the best baseline as the inner sequence model of the EM-Network, and thus the parameter for $\theta$ corresponded to that of the best baseline, such as data2vec and BiBERT. The size of the additional parameters $\rho$ was mainly determined by the vocabulary size. Since the EM-Network for WMT'14 dataset used about 52K vocabulary for the decoder, it required the relatively large parameters for $\rho$.
However, during the inference, \emph{only the sequence model was used to generate the prediction}. Since both the oracle encoder and fusion module were removed during the inference, the computational load of $\rho$ was only required for the training procedure.
Considering the performance improvements of the proposed approach, this computational load for training seems reasonable.

\section{Effect of $\alpha$ and $\lambda$}
\begin{figure}[h]
\centering
	\includegraphics[height=4.7cm]{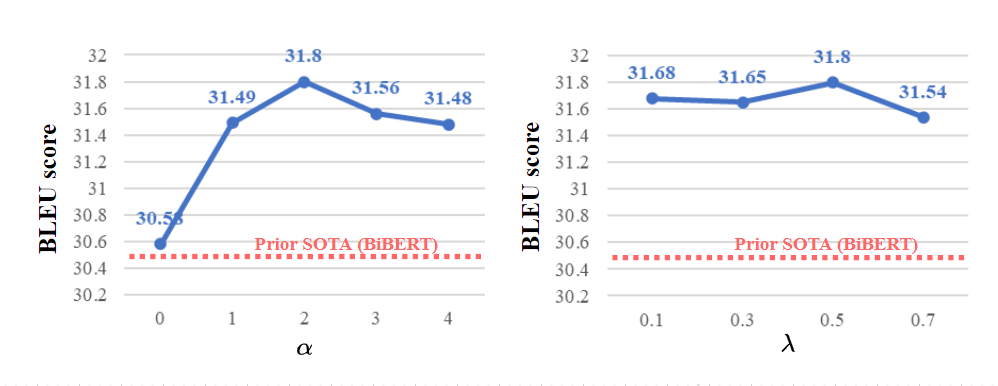}
	\caption{IWSLT'14 En-De performance with varying $\alpha$ and $\lambda$.}
	\label{ablation}
\end{figure}
When training the AED-based EM-Network, we considered two tunable parameters $\alpha$ in Eq. (\ref{train_loss}) and $\lambda$ in Eq. (\ref{ce_em}). The parameter $\alpha$ was used to balance the distillation loss $\mathcal{L}_{kd}$, and $\lambda$ was applied as the masking probability in the masking function $\mathcal{M}$.
We explored the effect of $\alpha$ and $\lambda$ on EM-Network performance, as shown in Figure \ref{ablation}.
Firstly, we evaluated the EM-Network on IWSLT'14 En-De translation while varying $\alpha$. 
$\alpha$ was selected from \{0, 1, 2, 3, 4\}, and $\lambda$ was set to 50 \%.
We observed that on IWSLT'14 En-De, the performance was stably better than the prior SOTA approach, and the best performance was obtained at $\alpha=2$.
Interestingly, the EM-Network with $\alpha=0$ was slightly better than the performance of the BiBERT, but the difference was negligible.
Secondly, we showed the effect of $\lambda$ on EM-Network performance.
In this case, $\alpha$ was set to 2, and we selected $\lambda$ from \{10\%, 30\%, 50\%, 70\%\}.
From the results, it is verified that performance on IWLST'14 En-De achieved the best result when
$\lambda=50\%$.

\section{Additional Visualization}
\label{ctc_align_add}

\begin{figure}[h]
\centering
\includegraphics[height=14cm]{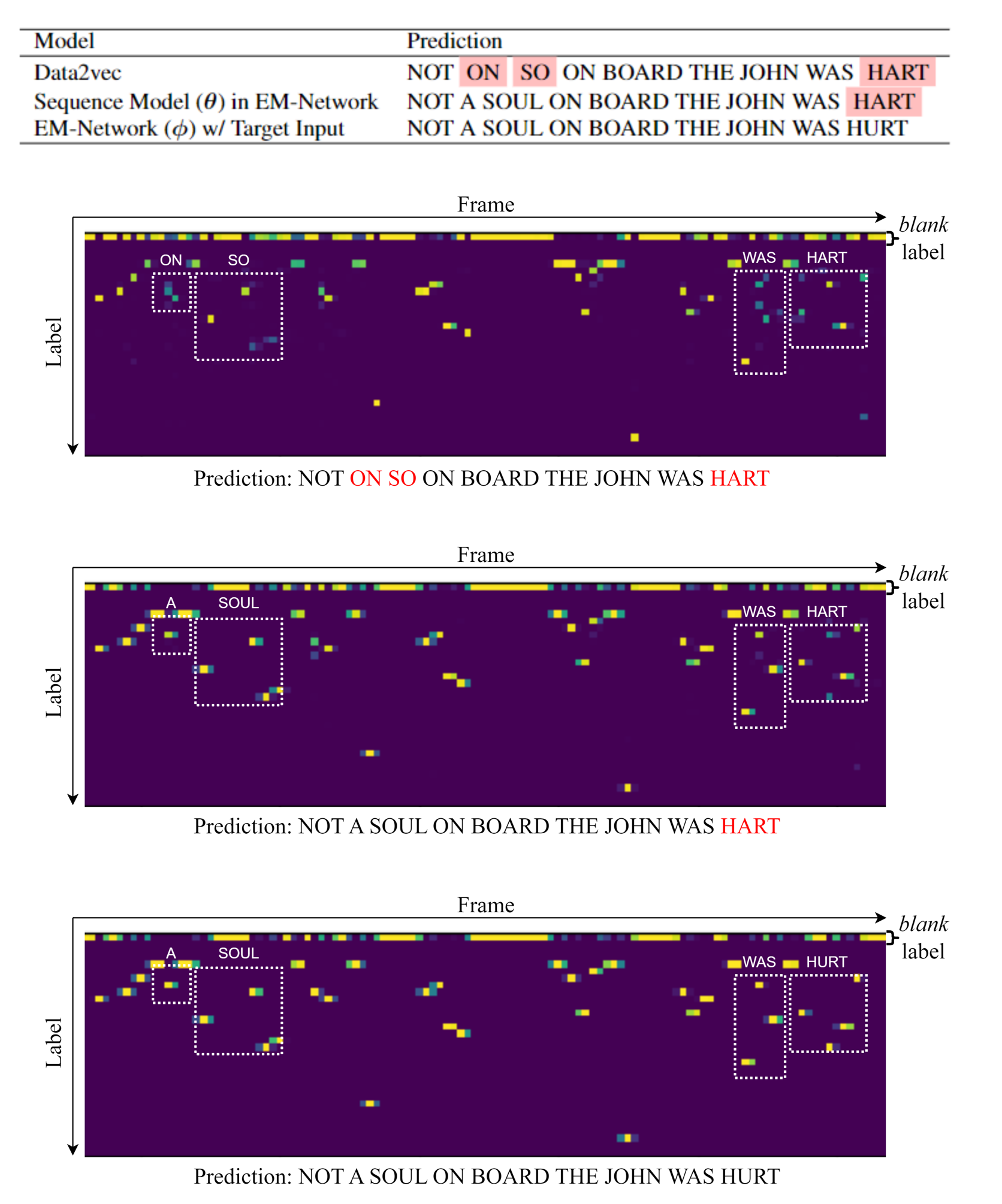}
\caption{Frame-wise label probability examples for utterance 2609-169640-0020 in LibriSpeech test-other dataset, where the target reference is ``NOT A SOUL ON BOARD THE JOHN WAS HURT".
Note that its argmax value corresponds to the predicted CTC alignment. The x-axis refers to acoustic frames, and the y-axis refers to
the character labels. The first label index represents the “blank" label in the CTC framework.}
\label{ctc_align_add_figure}
\end{figure}

We additionally compared the total frame-wise softmax outputs of the best baseline data2vec, our sequence model $\theta$ (student mode) in the EM-Network, and the EM-Network $\phi$ (teacher mode), as visualized in Figure \ref{ctc_align_add_figure}. The argmax value of the frame-wise label probability corresponds to the predicted CTC alignment.
The conventional data2vec made erroneous predictions with “ON SO" and “HART". 
When considering only the acoustic feature (speech voice), it is challenging to distinguish “ON SO"/“A SOUL" and “HART"/“HURT".
However, the EM-Network gave the correct prediction result with few irrelevant alignments, indicating that the proposed method effectively leveraged the target information to find the optimal CTC alignment.
By benefiting from the knowledge of the EM-Network, the proposed sequence model provided a more accurate prediction than the data2vec while it gave "A SOUL" instead of "ON SO".

By Eq. (\ref{lower_ctc}), we confirmed that minimizing $\mathcal{L}_{kd}$ maximizes a lower bound of the CTC model's log-likelihood. The lower bound is tight when achieving a low distillation loss between EM-Network and the sequence model, meaning a relatively large overlap between the predictions. 
 Due to the alignment-free property of the CTC framework, the conventional CTC models trained with the same settings, such as model architecture, training data, etc., can have different frame-level alignments, and this difference often leads to the convergence issue of the distillation \cite{tutornet:scheme}.
However, as shown in Figure \ref{ctc_align_add_figure}, 
 both EM-Network (teacher mode) and the sequence model (student mode) could produce similar frame-wise predictions (green boxes in Figure \ref{ctc_align_add_figure}), indicating that the proposed method was indeed able to perform the frame-wise distillation effectively and give a considerable alignment overlap, making a tight lower bound.

\section{Additional Experimental Results}
\paragraph{Offline Distillation with EM-Network.}
The proposed method was performed with online distillation, meaning the student and teacher were trained together. To verify the effect of online distillation, we transferred the EM-Network's knowledge with the offline distillation setup.
We trained the whole EM-Network with the CTC objective and extracted the knowledge from the pre-trained EM-Network. Then, we trained the student model (sequence model) using the extracted knowledge from the EM-Network. To fairly compare the results, we used SKD as the KD technique. The Conformer-CTC Small was adopted as the student baseline.

\begin{table*}[h]
\caption{Comparison of WER (\%) on LibriSpeech when greedy decoding was applied.}
\vskip 0.15in
\centering
\begin{small}
\begin{tabular}{cccccc}
\hline
Method &	KD	& {\begin{tabular}[c]{@{}c@{}}Additional \# of \\Params. for Teacher\end{tabular}}&	Student	 &test-clean&	test-other \\
\hline
{\begin{tabular}[c]{@{}c@{}}Student Baseline \\(Conformer-CTC Small)\end{tabular}}	& None	& None	& None& 	4.87&	12.05 \\
\hline
EM-Network& Offline	& 14 M	& Conformer-CTC Small & 	4.35 &	10.97 \\
\hline
EM-Network&	Self (Online)&	1 M	&Conformer-CTC Small&	4.29&	10.81\\
\hline
\label{wer_offline}
\end{tabular}
\end{small}
\end{table*}

As shown in Table \ref{wer_offline}, the EM-Network performed better with online self-distillation than offline KD. In contrast to offline distillation, where the knowledge was fixed, EM-Network with self-distillation could transfer knowledge more adaptively by updating the soft labels in each iteration, resulting in better performance improvement. From the results, it is verified that the EM-Network with online self-distillation was not only advantageous in terms of simplicity and computational efficiency but also performance improvement.

\paragraph{Application to Hybrid CTC/Attention Model.}
In the conventional hybrid CTC/attention architecture \cite{hybridctc}, the AED and CTC models share the encoder output. This enables the backpropagation of text information from the AED to the encoder, potentially benefiting the CTC model.
To further validate the effectiveness of distilling additional text information using the proposed framework, we applied the EM-Network approach to the hybrid CTC/attention Transformer model. For implementation, we used the ESPNet \cite{watanabe2018espnet} toolkit. When applying the proposed method, the mask probability in Eq. (\ref{ce_em}) was set to 0.5. The student Transformer model consisted of 3 encoder layers and 3 decoder layers. When training the student model, we used four Quadro RTX 8000 GPUs (each with 48GB of memory), and 50 epochs were spent for training. During the training, we did not apply the SpecAugment \cite{specaugment}. For the conventional teacher model, we used the pre-trained checkpoint provided by the ESPNet toolkit. During the inference, we averaged 5 chechpoints, following the original configuration of ESPNet. We used pre-trained Transformer LM, provided by the ESPNet. The beam size was set to 10.

\begin{table*}[h]
\caption{Comparison of WER (\%) on LibriSpeech using LM.}
\vskip 0.15in
\centering
\begin{small}
\begin{tabular}{cccccc}
\hline
Method &	{\begin{tabular}[c]{@{}c@{}}Additional \# of \\Params. for Teacher\end{tabular}} &	KD	 & test-clean &	test-other \\
\hline
Hybrid CTC/attention Transformer (Teacher for Standard KD) &	None&	None	&2.7&	5.9\\
\hline
Hybrid CTC/attention Transformer (Student Baseline)	&None&	None&	4.1&	9.4 \\
\hline
Standard KD & 	78M	&Offline	&3.9	&9.2\\
\hline
Ours, EM-Network &	7M&	Self&	3.6	&8.8\\
\hline
\label{wer_ctc_attention}
\end{tabular}
\end{small}
\end{table*}

From the results, we confirmed that the EM-Network still performed well for the AED-based ASR model, like the EM-Network was effective for the AED-based MT models in our experiments. This implies that utilizing more global text information via the proposed framework was supportive in training the AED-based ASR model.

\paragraph{Comparison with Text Injection ASR Techniques.}
There have been several studies that attempted to jointly learn shared speech and text representation \cite{speechlm, tessp, token2vec, text-injection1, text-injection2}, especially for self-supervised learning settings.
Note that previous text injection ASR models jointly learned shared speech and text representation in the self-supervised setting, while our method considered supervised learning. We compared the proposed method with recent text injection techniques, including SpeechLM \cite{speechlm}, TESSP \cite{tessp}, and Token2vec \cite{token2vec}. For SpeechLM and TESSP, they used the paired data, which was used to learn the alignment between the speech frames and the text input token, during the pre-training stage. For the SpeechLM, we used the hidden-unit tokenizer using HuBERT \cite{hubert}. All the SSL models were fine-tuned using 100 hours of LibriSpeech. When applying our method, we took the pre-trained data2vec \cite{data2vec} model as our backbone and fine-tuned it using our method with 100 hours of LibriSpeech. To fairly compare the results, we employed one linear layer to perform the CTC during the fine-tuning, and the SSL models were based on Base setting with 12 Transformer encoder layers.

\begin{table*}[h]
\caption{Comparison of WER (\%) on LibriSpeech.}
\vskip 0.15in
\centering
\begin{small}
\begin{tabular}{ccccc}
\hline

Model&	Pre-training Data &	Fine-tuning Data	&test-clean	&test-other \\
\hline
Wav2vec 2.0&	Libri960h(unpaired)	&Libri100h&	6.1&	13.3\\
\hline
HuBERT&	Libri960h(unpaired)&	Libri100h&	6.3	&13.2\\
\hline
WavLM	&Libri960h(unpaired)&	Libri100h&	5.7	&12.0\\
\hline
ILS-SSL&	Libri960h(unpaired)&	Libri100h	&4.7&	10.1\\
\hline
Data2vec	&Libri960h(unpaired)	&Libri100h&	4.5&	9.8\\
\hline
TESSP	&Libri960h(unpaired)+Libri100h(paired)+Libri LM corpus(text)&	Libri100h&	5.1	&10.9\\
\hline
Token2vec	&Libri960h(unpaired)+ Libri LM corpus(text)&	Libri100h&	5.1	&11.8\\
\hline
SpeechLM	&Libri960h(unpaired)+Libri100h(paired)+Libri LM corpus(text)&	Libri100h&	4.3&	10.0\\
\hline
EM-Network (Ours)	&Libri960h(unpaired)&	Libri100h&	4.1	&9.8\\
\hline
\label{text-injection}
\end{tabular}
\end{small}
\end{table*}

As shown in Table \ref{text-injection}, it is verified that the EM-Network performed better compared to the previous text injection methods. 
Different from previous text injection methods, our proposed approach combined text injection and distillation during the fine-tuning stage. 
From the results, it is confirmed that the EM-Network achieved the best performance without requiring additional speech-text paired dataset or large text corpora during the pre-training phase.
This suggests that the proposed method could serve as an effective and efficient alternative to conventional text injection methods.
Since we used train-clean-100 (LibriSpeech 100 hours) during the fine-tuning, the proposed method showed no significant improvement on the test-other dataset.

\section{Difference from Original CTC Backward}
The gradient for the original CTC is as follows:
\begin{align}
\frac{\partial \mathcal{L}_{ctc}}{\partial u_{k}^{t}}=p(z^t=k|x; \theta)-\sigma_{ctc}(k,t)
\end{align}
where $u_{k}^{t}$ and $\sigma_{ctc}(k,t)$ represent the logit and posterior probability of the $k$-th phoneme at frame $t$, respectively. The term $\sigma_{ctc}(k,t)$ is calculated using the forward-backward algorithm. The conventional CTC produces a valid alignment $\sigma_{ctc}(k,t)$ through the forward-backward calculation on the transcription. However, CTC models often converge to sub-optimal alignment distributions \cite{ctc-peaky,ctc-peaky2,ctc-peaky3}.

In the EM-Network framework, the gradient for the proposed distillation between the sequence model and the EM-Network is given by
\begin{align}
\frac{\partial \mathcal{L}_{kd}}{\partial u_{k}^{t}}=p(z^t=k|x; \theta)-p(z^t=k|x,y;\phi)
\end{align}
where $p(z^t=k|x,y;\phi)$ is the $k$-th phoneme outputs at frame $t$ of the EM-Network model. Different from the original backward that uses the target in the forward-backward calculation (for generating correct alignment), we apply the auxiliary modules, including the oracle encoder and fusion module, on top of the original sequence model $\theta$ (where $\theta$ $\subset$ $\phi$).
The oracle encoder consumes the target input $y$ and generates the oracle guidance.
Then, the fusion module performs a fusion of the CTC model's output and the oracle guidance, refining the original CTC model's output based on the target-side information.
Now, the output of the EM-Network ($\phi$) is more accurate than that of the original CTC model ($\theta$), which can be seen in Figure \ref{loss} (Section \ref{analysis_sec}). It means that the output of the EM-Network is the supportive knowledge for training the CTC model.
The probability $p(z^{t}=k|x, y;\phi)$ is derived from the EM-Network, and the CTC model is optimized to imitate it, no matter whether it is correct or wrong compared with the transcription. The probability $p(z^t=k|x,y;\phi)$ of the EM-Network acts like the soft label, which is the key difference from the original backward process. From the KD perspective, the performance improvement by using the soft label seems reasonable.
In addition to the original CTC backward, the proposed $p(z^t=k|x,y;\phi)$ can provide additional alignment information in training the CTC alignment. 
Also, the quality of $\sigma_{ctc}(k,t)$ is the bottleneck of the convergence speed and performance. 
Thus, properly optimizing the parameter $\theta$ is an important factor for the CTC model's performance. 
If we can effectively model the posterior $p(z^t=k|x,y;\phi)$, the CTC model is able to produce more accurate alignment compared to the original CTC backward.
Empirically, we observe that the EM-Network gives the promising alignment based on its posterior $p(z^t=k|x,y;\phi)$, and the parameter $\theta$ can be optimized properly by using the EM-Network's supportive knowledge.

\section{Limitations and Societal Impacts}
\paragraph{Limitations.} This work offers a powerful self-distillation method in the context of the sequence learning tasks and leaves other domains, such as image classification, for future work. 
For the CTC framework, the many-to-one relationship between the CTC alignment $z$ (many) and the target $y$ (one) is the key to training the EM-Network. Intuitively, predicting the latent alignment $z$ (many) from the target input $y$ (one) should be difficult since many possible paths are compatible with $y$. We can easily apply the EM-Network to the CTC-based model since the CTC algorithm inherently prohibits the trivial solution by employing the many-to-one mapping function. In the case of the AED-based EM-Network, we use random masking to prevent the aforementioned trivial solution.
Even though the EM-Network is capable of modeling a wide range of the seq2seq networks, it is difficult to directly apply our framework to non-sequence domains, like image classification and language understanding tasks.
Also, the oracle guidance derived from the target could limit the usage on unsupervised data. 
To apply our approach to the unsupervised setting, the model's prediction can be utilized instead of the target input, like \cite{imputer,mask-ctc}, and we will test it in the future.

\paragraph{Potential Negative Societal Impact.}
Recently, the distillation framework becomes more important in a real-world deployment.
While the proposed method gave comprehensive and convincing results for various sequence tasks, additional analysis and study are necessary before applying it in practice.
For example, the EM-Network model needs both the source input $x$ and the target input $y$ for training.
The data like speech or text may contain private information. In other words, the EM-Network, trained on such sensitive data, can cause privacy leakage.

%%%%%%%%%%%%%%%%%%%%%%%%%%%%%%%%%%%%%%%%%%%%%%%%%%%%%%%%%%%%%%%%%%%%%%%%%%%%%%%
%%%%%%%%%%%%%%%%%%%%%%%%%%%%%%%%%%%%%%%%%%%%%%%%%%%%%%%%%%%%%%%%%%%%%%%%%%%%%%%

\end{document}